\pdfoutput=1

\documentclass{article}
\usepackage{ijcai25}

\usepackage{times}
\usepackage{soul}
\usepackage{url}
\usepackage[hidelinks]{hyperref}
\usepackage[utf8]{inputenc}
\usepackage[small]{caption}
\usepackage{graphicx}
\usepackage{amsmath}
\usepackage{amsthm}
\usepackage{booktabs}
\usepackage{algorithm}
\usepackage{algorithmic}
\usepackage[switch]{lineno}
\usepackage{multirow}
\usepackage{bm}

\title{HFGCN:Hypergraph Fusion Graph Convolutional Networks for \\
Skeleton-Based Action Recognition}

\author{
Pengcheng Dong$^1$\and
Wenbo Wan$^1$\and
Huaxiang Zhang$^1$\and
Shuai Li$^2$\and
Sujuan Hou$^1$\and
Jiande Sun$^1$$^*$\\
\affiliations
$^1$School of Information Science and Engineering, Shandong Normal University, China\\
$^2$School of Control Science and Engineering, Shandong University, China\\
\emails
2022317067@stu.sdnu.edu.cn,
wanwenbo@sdnu.edu.cn,
huaxzhang@163.com,
shuaili@sdu.edu.cn,
sujuanhou@sdnu.edu.cn,
jiandesun@hotmail.com
}

\begin{document}

\maketitle
\renewcommand{\thefootnote}{}
\footnotetext[1]{\textsuperscript{*}Corresponding author.}

\begin{abstract}
In recent years, action recognition has received much attention and wide application due to its important role in video understanding.
Most of the researches on action recognition methods focused on improving the performance via various deep learning methods rather than the classification of skeleton points.
The topological modeling between skeleton points and body parts was seldom considered.
Although some studies have used a data-driven approach to classify the topology of the skeleton point, the nature of the skeleton point in terms of kinematics has not been taken into consideration.
Therefore, in this paper, we draw on the theory of kinematics to adapt the topological relations of the skeleton point and propose a topological relation classification based on body parts and distance from core of body.
To synthesize these topological relations for action recognition, we propose a novel Hypergraph Fusion Graph Convolutional Network (HFGCN).
In particular, the proposed model is able to focus on the human skeleton points and the different body parts simultaneously, and thus construct the topology, which improves the recognition accuracy obviously. We use a hypergraph to represent the categorical relationships of these skeleton points and incorporate the hypergraph into a graph convolution network to model the higher-order relationships among the skeleton points and enhance the feature representation of the network.
In addition, our proposed hypergraph attention module and hypergraph graph convolution module optimize topology modeling in temporal and channel dimensions, respectively, to further enhance the feature representation of the network.
We conducted extensive experiments on three widely used datasets,i.e., NTU RGB+D, NTU RGB+D 120, and NW-UCLA.The results validate that our proposed method can achieve the best performance when compared with the state-of-the-art skeleton-based methods.

\end{abstract}

\section{Introduction}

Human action recognition holds great significance in daily life. It allows computers to perceive and interpret human actions and behaviors.
This technology has broad applications across various domains such as video surveillance, sports training, medical rehabilitation, and human-computer interaction\cite{poppe2010survey,yang2019gesture,pareek2021survey}. In recent years, the increasing emphasis on intelligent sports analysis has further broadened the scope of human action recognition within the sports domain. Applications include real-time action evaluation, posture correction, athlete performance analysis\cite{host2022overview,wu2022survey}. These applications are particularly beneficial in sports science, where the accurate detection and analysis of actions can provide valuable insights for training and rehabilitation. However, sports applications pose distinct challenges compared to other fields. They require both high detection accuracy and rapid response times to deliver real-time feedback.

One significant advantage of skeleton-based action recognition over traditional RGB-based methods is that it  can utilize human skeleton points, which are extracted by sensors like Kinect\cite{bawa2021review} or reliable human pose estimation algorithms\cite{song2021human} for action recognition. 
Unlike RGB-based methods, which are susceptible to lighting change and background noise, skeleton-based methods focus on the geometric relationships between human skeleton points, which makes them more robust to environmental variations. Skeleton-based methods are also more computationally efficient, as they operate on a smaller set of human skeleton points. 
This efficiency makes it well-suited for real-time applications, especially in the sports domain, where both speed and accuracy are essential.

Early methods for action recognition relied heavily on manually designed features, such as handcrafted descriptors and trajectories\cite{xia2012view,vemulapalli2014human}. 
These methods achieved acceptable performance in controlled settings. However, manually designed features are unable to fully capture the inherent complexity of human action, particularly for long-duration (as all actions are dynamic in nature) and complex actions. These methods failed to adapt to diverse actions. As a result, there was a noticeable drop in accuracy. With the rise of deep learning, models based on Recurrent Neural Networks (RNNs)\cite{du2015hierarchical,singh2016multi,wang2017modeling} and Convolutional Neural Networks (CNNs)\cite{simonyan2014two,tran2015learning} have been widely adopted for action recognition tasks. These methods have demonstrated improved performance by automatically learning feature representations from data. However, these methods often struggle to effectively model the spatial and temporal dependencies in human skeleton data.

Recently, Graph convolutional Networks (GCNs)\cite{kipf2017semisupervised} have emerged as a mainstream method for skeleton-based action recognition due to their ability to model spatial and temporal dependencies using graph data. In the skeleton-based data, human joints are treated as graph vertices,, and the connections between joints are represented as edges in a graph. GCNs have demonstrated their effectiveness in modeling these topologies and learning hierarchical representations from joint-based features. The mainstream approach in this field represents skeleton sequences as spatiotemporal graph structures. It uses GCNs to improve the learning of topological features for action recognition. The Spatial-Temporal Graph Convolutional Network (ST-GCN) \cite{yan2018spatial} is the pioneering work that applies GCNs to skeleton-based action recognition. ST-GCN collects information from adjacent joints based on the natural physical connections of the human body. It constructs a topological graph that mirrors these connections.
However, the manually defined topological graph restricts the capacity of model to capture non-physical relationships between joints. This limitation reduces the performance of model and results in suboptimal performance on complex action recognition tasks. To overcome this limitation, several studies\cite{shi2019two,li2019actional,liu2020disentangling,chen2021channel,chi2022infogcn} have explored methods to enable the model to adaptively learn the correlations between different joint pairs.

Despite progress in optimizing topological graphs, GCN-based methods still face two major limitations that impact their performance. Firstly, these methods primarily emphasize learning pairwise relationships between individual human skeleton points. They overlook the significance of skeleton point groups in representing human actions. Human skeleton points of the same type play similar physical roles in certain actions  and should be treated as skeleton point groups rather than isolated skeleton points. For example, the elbow and knee joints have similar degrees of motion during running or jumping, and should be modeled as a group to better understand the physical dynamics of the action. By modeling these skeleton point groups and their interrelationships, recognition accuracy can be significantly improved. One promising solution is the use of hypergraphs \cite{bretto2013hypergraph,feng2019hypergraph} to represent high-order relationships between human skeleton points. A hypergraph is a generalization of a traditional graph in which a hyperedge can connect multiple vertices. This allows the modeling of higher-order relationships.
This is particularly valuable in human action recognition, where skeleton point groups, rather than pairwise connections, play a more significant role in certain actions. We propose two novel hypergraph constructions. Specifically, we introduce two graph topology structures: one that groups skeleton points by human body parts (e.g., arms, legs, torso), and another based on their spatial distance from the core of the human body. This results in a hypergraph topology that emphasizes the relationships between various skeleton point groups. In addition, we propose a Hypergraph Attention Module (HAM) to process both the original skeleton data and the grouped skeleton data, thereby obtaining the final hypergraph topology. It facilitates more effective modeling of the human skeleton structure and enhances the model performance in action recognition.

Secondly, current GCN-based methods are unable to adequately represent dynamic information in the topological graphs. Existing methods typically use a fixed topology across all frames of a sample, which fails to capture the temporal variations in the relationships between human skeleton points. In real-world action recognition tasks, these relationships can change over time.  It is critical to capture such dynamic variations for accurate action classification. To address this issue, we introduce modules that operate across both temporal and spatial channels to capture the evolving relationships between skeleton points and joint groups. Specifically, we propose the HAM and the Hypergraph Convolution Module (HGCM), which optimize the topological representations of skeleton data across different temporal frames and spatial channels. By capturing the temporal dependencies and incorporating dynamic topological changes, these modules enhance the ability to recognize action of model.

The key contributions of our proposed hypergraph fusion graph convolutional network (HFGCN) can be summarized as follows:

First, we group the human skeleton points into distinct human body parts and construct a classification hypergraph matrix based on these groupings. This novel approach allows the model to highlight the correlation between the skeleton points and the human body parts using various classification methods, thereby improving the capacity for topological modeling of model.

Second, we propose a HAM, that capture both pairwise relationships between different skeleton points as well as higher-order relationships between skeleton points and human body parts, and enhances topological modeling in the temporal dimension.

Finally, we develop a HGCM, that serves as a channel topology modeling branch. It optimizes the topology modeling in the feature channel dimension and aggregates the topologies obtained from HAM to enhance the topology modeling capability of the model.

\section{Releated Work}

\subsection{GCN-Based Action Recognition}
GCN-based action recognition methods are typically categorized into two primary approaches: spectral domain-based methods and spatial domain-based methods. Spectral domain-based methods extract graph properties using the features of the graph's Laplacian matrix, while spatial domain-based methods apply graph convolution to directly extract features from topological graphs that represent the relational connectivity among graph vertices. Due to their superior flexibility and generalization ability, most GCN-based action recognition schemes adopt spatial-domain methods.

Current GCN-based methods primarily rely on message passing, which enables the model to recognize actions using an attention matrix. This matrix captures semantic relationships, context, and connectivity between human skeleton points. Yan et al. \cite{yan2018spatial} first introduced spatiotemporal graph convolution, which uses a predefined adjacency matrix based on prior knowledge to effectively learn spatiotemporal features from skeleton data. However, the adjacency matrix in ST-GCN is fixed and remains unoptimized during training, which limits the model's performance. Ding et al. \cite{ding2019attention} and Shi et al. \cite{shi2019two} significantly improved recognition performance by using adaptive learning methods to optimize topological modeling. Beyond adaptive learning, Cheng et al. \cite{cheng2020skeleton} introduced a global shift mechanism that allows the receptive field of each node to cover the entire graph. Chen et al. \cite{chen2021channel} further enhanced adaptivity and overall performance by introducing the Channel-wise Topology Refinement Graph Convolution (CTRGC) module, which models different topological matrices for different channels. However, these methods generally do not account for the dynamic evolution of spatial topology over time, a limitation that impedes model performance. Chi et al. \cite{chi2022infogcn} addressed this issue by incorporating Transformer-based architectures, which allow the model to adaptively learn and construct topology based on latent information. Liu et al. \cite{liu2023temporal} performed optimization of topological modeling over the temporal dimension. Nevertheless, these approaches often neglect the relationships between individual skeleton points and groups of skeleton points, which are crucial for accurate action recognition.

\subsection{Hypergraph Learning}

Instead of a edge that connects two vertices, a hypergraph has hyperedges that can connect any number of vertices. In skeleton-based action recognition, the higher-order correlations across multiple skeleton points are more aptly captured by the use of hyperedges in hypergraphs.
Most existing hypergraph models primarily focus on hypergraph construction and learning. For example, Hyper-GNN \cite{hao2021hypergraph} selects specific joint points as hyperpoints and uses sparse representation techniques to construct the hypergraph. However, identifying hyperpoints and calculating the reconstruction coefficients is computationally demanding. HyperGCN \cite{chen2022informed} utilizes human pose skeletons and informed visual patches for multi-modal feature learning, incorporating visual vertices into hypergraph edges to construct the hypergraph.
Zhou et al. \cite{zhou2022hypergraph} dynamically optimize the hypergraph topology through a self-attention mechanism combined with hypergraph learning.
However, while the hypergraph construction is data-driven, it is not specifically designed with human motion in mind. This limits its ability to capture motion-specific relationships.
\begin{figure*}[t]
    \centering
    \includegraphics[width=0.8\linewidth]{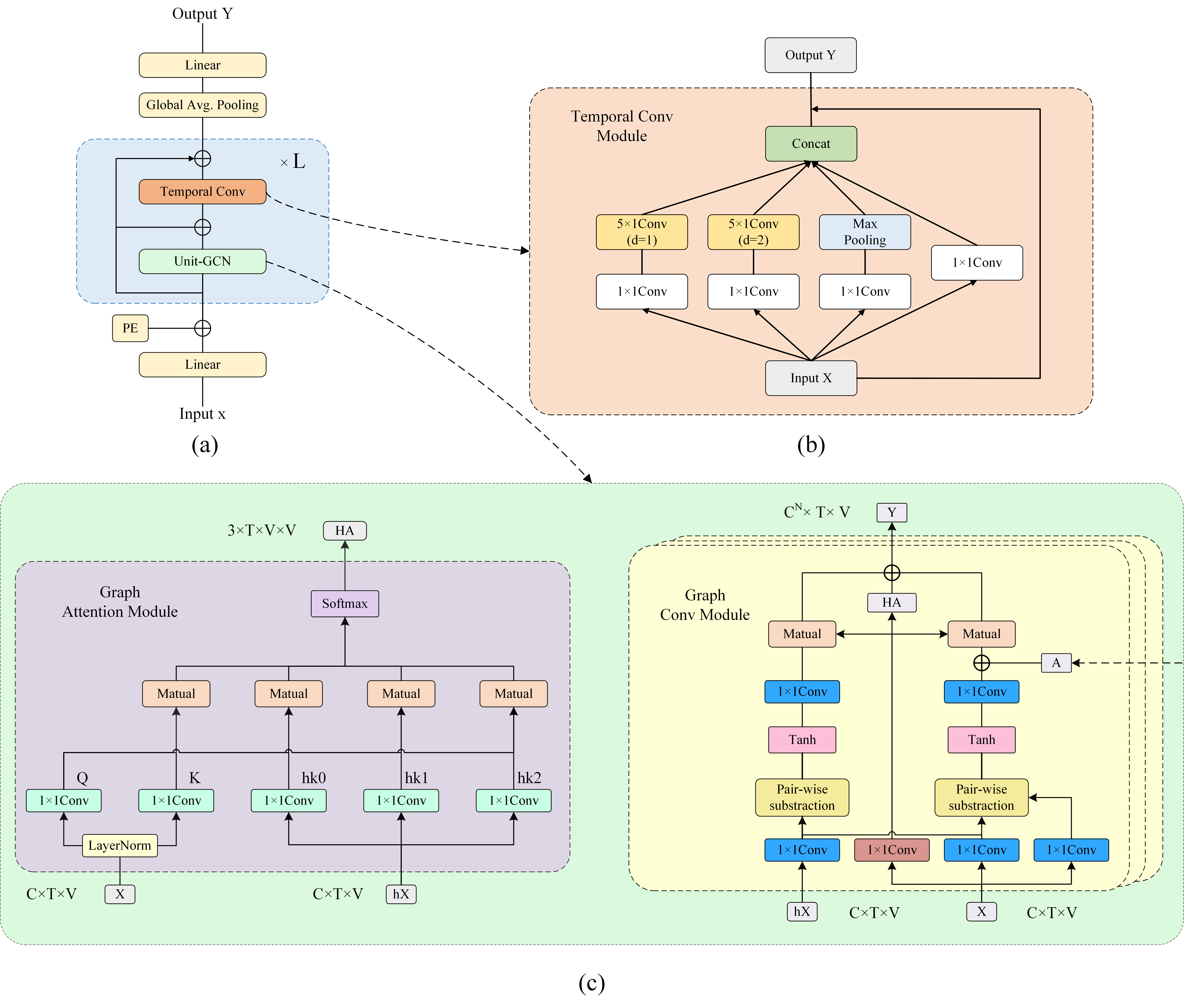}
    \caption{The illustration of our proposed HFGCN architecture, (a) is the overall architecture of HFGCN, (b) is the temporal convolution module, and (c) is the module in the Unit-GCN, which are the hypergraph attention module and the hypergraph convolution module respectively.}
    \label{fig:1}
\end{figure*}

\section{Method}
In this section, we first present the relevant formulas and mathematical definitions of graph convolution, and then introduce our Hypergraph Fusion Convolutional Network (HFGCN). Subsequently, we detail the construction of the hypergraph topology, followed by the introduction of our HAM and HGCM.

\subsection{Graph Convolution}
The human skeleton sequence naturally forms a topological graph, where the joints correspond to the vertices and the bones to the edges. This graph can be defined as $\bm{G=\{V,E\}}$, where $\bm{V=\{v_1,v_2,...v_N\}}$ represents the set of $N$ joints, and $\bm{E}$ denotes the set of bones in the skeleton sequence. For 3D skeleton sequence data, each joint $\bm{v_i}$ is represented as $\bm{v_i = \{x_i, y_i, z_i\}}$, where $\bm{x_i}$, $\bm{y_i}$, and $\bm{z_i}$ are the coordinates of joint $\bm{v_i}$ in 3D space.

Let $\bm{x_j}$ and $\bm{y_j}$ denote the input and output data, respectively. The standard spatial graph convolution operation, commonly used in skeleton-based action recognition, is formulated as:
\begin{equation}
\label{deqn1}
\bm{y_j = \sigma \left( \sum_{\boldsymbol{v}_i \in N(\boldsymbol{v}_j)} a_{ij} x_j W \right)}
\end{equation}
where $\sigma(\cdot)$ denotes an activation function, $N(\cdot)$ represents the set of neighboring joints, $W$ is the weight matrix used for feature transformation, and $a_{ij}$ is the $(i,j)$ entry of the normalized adjacency matrix $A\in R^{N\times N}$ that signifies the correlation strength between $v_i$ and $v_j$.

\subsection{Hypergraph Topology Construction} 
Recently, many methods \cite{thakkar2018part,huang2020part,song2020stronger} have paid attention to the importance of skeleton point groups in action recognition. These methods focus on learning which body parts are most relevant for specific actions. This helps the model classify actions more effectively. Building on this idea, we explored the impact of using different skeleton point classification strategies and their fusion.
In competitive sports, there is a well-known event-group training theory, which organizes items based on their shared attributes and training requirements. Items with similar characteristics are grouped together to identify common features and development patterns. Similarly, different body movements often share characteristics, where the same skeleton points may play roles in multiple actions. Inspired by this, we designed three different hypergraph topologies to help the model better perform action recognition.

\begin{figure}
    \centering
    \includegraphics[width=\linewidth]{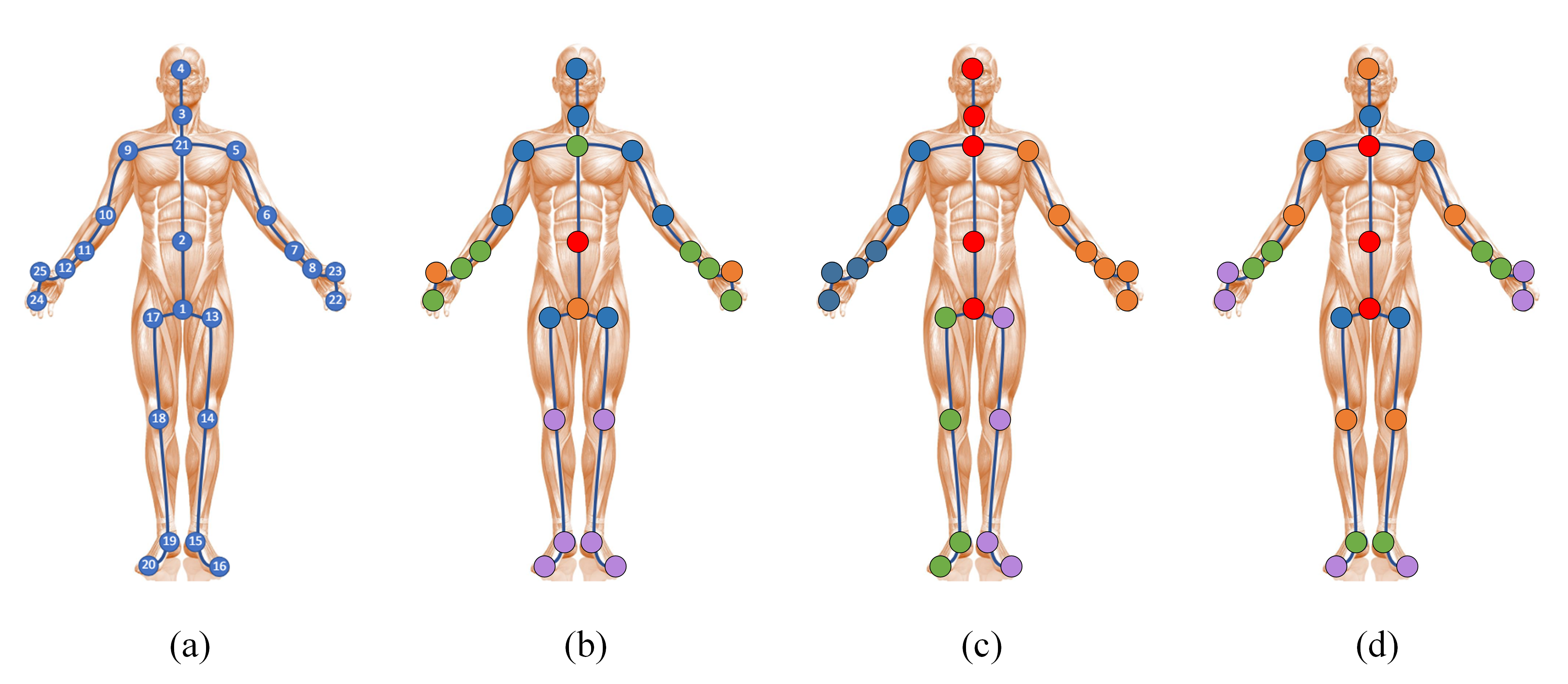}
    \caption{The illustration of the hypergraph topology construction. (a) is the skeleton topology, (b), (c) and (d) are the hypergraph topologies we constructed.}
    \label{fig:2}
\end{figure}

The proposed hypergraph topologies are illustrated in Fig \ref{fig:2}. Firstly, we adopted the classification strategy from \cite{zhou2022hypergraph}. Secondly, drawing on insights from sports science and biomechanics, we classified the body into distinct body parts, separating the arms, legs, and torso, with the head grouped with the torso due to its limited range of movement. Most actions are driven by large muscle groups that activate smaller muscle groups. The arms, legs, and torso are typically the primary focus.
Finally, we introduced a classification strategy based on the distance of skeleton points from the core of body, reflects the principle that large, proximal muscle groups generate movement, while distal endpoints move in response.

With these three classification strategies, we then construct a hypergraph for each strategy and incorporate them into the input data.
These hypergraphs lays a solid foundation for effective topological modeling in subsequent network learning.

\subsection{Hypergraph Attention Module}
In the hypergraph topology we designed, different skeleton points are grouped into distinct categories. This structure highlights the relationships between individual skeleton points and hypergraph categories, enhancing the representation of structural dependencies.
Existing methods construct topological graphs for samples, but they fail to adequately capture the relationships between skeleton points and body parts. As a result, they are unable to effectively account for skeleton data across different frames and action types.
To address this limitation, we developed a HAM to capture the high-order correlations between skeleton points and body parts. This module constructs a multi-frame and multi-branch dynamic hypergraph topology, significantly enhancing the expressiveness of the GCN.

Our proposed model architecture is shown in Fig \ref{fig:1} (c). Here, $X$ represents the original input feature, while $hX$ is obtained by multiplying the three hypergraph topological structures with the original feature $X$. This process integrates hypergraph information into the feature $X$. The shape of $hX$ is $(3, B, C, T, V)$, where $B$ is the batch size, $C$ is the number of channels, $T$ is the number of frames, and $V$ represents the number of joints (25 in this case). 
We input the feature $X$ into the HAM, where it first passes through a $1 \times 1$ convolution to generate queries $q$ and keys $k$. Simultaneously, the input $hX$ is divided into three distinct groups of features, each processed through a $1\times 1$ convolution to create values $hk_0, hk_1$, and $hk_2$. These are then multiplied to compute the point-pair correlations and point-group correlations under different hypergraph topologies. The resulting features are aggregated and processed using a softmax activation function to generate the final topology $HA$ of shape $(3,T,V,V)$, which establishes the relationships between skeleton points and body parts.

Additionally, our HAM incorporates the optimization of the topological structure across different frames. Unlike methods that employ a fixed topology for all frames, our HAM preserves the temporal dimension of the input. This allows our model to adapt the topology for each frame individually, thereby enhancing the feature representation capability of the model. To summarize, our HAM is defined as follows:
\begin{equation}
    HA = \text{Softmax}(q \cdot k + q \cdot hk_0 + q \cdot hk_1 + q \cdot hk_2)
\end{equation}

\subsection{Hypergraph Convolution Module}
To construct topological structures across different spatial channels, we build upon the work of \cite{chen2021channel} and incorporate hypergraph topology to develop a channel-level improved topology module. Fig. 1(c) shows our proposed Hypergraph Convolution Module (HGCM), which extracts the correlations between skeleton points as well as the relationships between skeleton point groups.
The proposed HGCM consists of three branches. The right branch is similar to the CTR-GC module in \cite{chen2021channel}. It takes the input features $X$, computes the distances between skeleton points in the channel dimension, and applies the tanh activation function to endow the nonlinear relationships within the channel-specific topological structure of the skeleton points. The physical connection topology of the skeleton points $A$ is combined with the optimized topology from the right branch, which is then multiplied by the standard feature $X$ to yield the output of the right branch.
The left branch computes the distance between the hypergraph feature and the standard feature along the channel dimension, applies a nonlinear transformation to capture the topological relationships between skeleton points and skeleton point groups, and then multiplies the result by the standard feature $X$ to generate the output of the left branch.
The middle branch multiplies the topology obtained from the HAM module with the standard feature $X$ to produce the output feature.

This branch primarily serves to transfer the topology optimized by the HAM module to the HGCM. This process ensures seamless integration, enabling effective fusion of the final feature representation. Finally, the outputs from the three branches are added and fused to obtain the final skeleton topology, achieving channel-level optimization of the skeleton topology. The overall improved topology in HGCM can be summarized as follows:

\begin{equation}
\begin{aligned}
    Y &= \delta(X) \cdot(\tanh(\varphi(X) - \psi(X)) + A)\\
      & \delta(X) \cdot \tanh(\varphi(X) - \xi(X)) + \delta(X) \cdot HA
\end{aligned}
\end{equation}
where $\varphi(\cdot)$, $\psi(\cdot)$, $\xi(\cdot)$ and $\delta(\cdot)$ represent different 1$\times$1 convolution operations, and $Y$ denotes the final output.

\subsection{Temporal Conv Module}
To model the temporal correlations of human pose, we incorporate the Multi-Scale Temporal Convolution (MS-TC) module \cite{liu2020disentangling,chen2021channel} into our final model. The architecture of this module is illustrated in Fig \ref{fig:1}(b). 
It consists of three convolution branches, each starting with a $1 \times 1$ convolution to reduce the channel dimension. This is followed by varying combinations of kernel sizes and dilations.
The outputs from these convolution branches are concatenated to form the final temporal feature representation.

\begin{table*}[t]
\centering
\caption{TOP-1 accuracy (\%) comparison with state-of-the-art methods on NTU RGB+D and NTU RGB+D 120 datasets. “-” indicates experimental results not provided in the references. Bold indicate the best results for each dataset. J stands for Joint, B stands for Bone, JM and BM stand for Joint motion and Bone motion.\label{tab:table1}}
\begin{tabular}{cccccccc}
\hline
\multirow{2}{*}{Methods} &\multirow{2}{*}{Publication} &\multirow{2}{*}{Modalities} &\multicolumn{2}{c}{NTU RGB+D} &\multicolumn{2}{c}{NTU RGB+D 120} & \multirow{2}{*}{NW-UCLA}\\ \cline{4-7} 

         &   & &X-Sub(\%) & X-View(\%) & X-Sub(\%) & X-Set(\%) & \\
\hline
ST-GCN                 & AAAI18  &J         & 81.5  & 88	   & 71.7	 & 72.2  & -     \\
AS-GCN          	     & CVPR19  &J         & 86.8  & 94.2   & 78.2	 & 77.7	 &-      \\
HFGCN(ours)              &         &J         & \textbf{90.9}  &\textbf{95.9}	   &\textbf{86.0}	 &\textbf{88.2}   &-\\
\hline
2s-AGCN                  & CVPR19  &J+B       & 88.5  & 95.1   & 82.9	 & 84.9  &-	    \\
HyperGCN &IPM22 &J+B &91.5 & 96.6 & 89.2 &90.2 &- \\
HFGCN(ours)                  &         &J+B       & \textbf{92.7}  &\textbf{96.8}	   &\textbf{89.5}	 &\textbf{91.1}   &-\\
\hline
ShiftGCN     & CVPR20  &J+B+JM+BM & 90.7  & 96.5   & 85.9	 & 87.6	 & 94.4 \\
MS-G3D	     & CVPR20&J+B+JM+BM 	& 91.5	& 96.2	 & 86.9	 & 88.4  &-      \\ 	
MST-GCN		 & AAAI21&J+B+JM+BM 	& 91.5	& 96.6	 & 87.5	 & 88.8	 &-      \\
CTR-GCN		 & ICCV21&J+B+JM+BM 	& 92.4	& 96.8	 & 88.9	 & 90.6	 & 96.5 \\
Hyper-GNN &TIP21 &J+B+JM+BM & 89.5 & 95.7 &- &- &- \\
EifficientGCN-B4 	    & TPAMI22&J+B+JM+BM 	& 92.1	& 96.1	 & 88.7	 & 88.9	 &-      \\
Hyperformer 	 & Arxiv22&J+B+JM+BM 	& 92.6	& 96.5	 & 89.9	 & 91.2  &-	    \\
InfoGCN	     & CVPR22&J+B+JM+BM 	& 92.7	& 96.9	 & 89.4	 & 90.7	 & \textbf{96.6} \\
ML-STGNet   & TIP23&J+B+JM+BM 	& 91.9	& 96.2	 & 88.6	 & 90	 &-      \\
FR-Head	                 & CVPR23&J+B+JM+BM 	& 92.8	& 96.8	 & 89.5	 & 90.9	 &-      \\   
TD-GCN    & TMM23&J+B+JM+BM 	& 92.8	& 96.8   &- 		 &-       &- 	    \\
MGCF-Net    & PR24&J+B+JM+BM 	& 92.7	& 96.8   & 88.7		 & 90.4     & -	    \\
\hline
HFGCN(ours) &       &J+B+JM+BM   & \textbf{93.1}	& \textbf{97.1}	 & \textbf{90.1} & \textbf{91.5} & 96.1 \\
\hline
\end{tabular}
\end{table*}

\begin{table}[t]
\centering
\caption{Comparison of TOP-1 accuracy of multi-modal ensemble on NTU RGB+D and NTU RGB+D 120. J stands for Joint, B stands for Bone, JM and BM stand for Joint motion and Bone motion.\label{tab:table2}}
\resizebox{1.0\linewidth}{!}{
\begin{tabular}{ccccc}
\hline
\multirow{2}{*}{Method} & \multicolumn{2}{c}{NTU RGB+D 60} &  \multicolumn{2}{c}{NTU RGB+D 120}\\ \cline{2-3} \cline{4-5}
  & X-Sub(\%) & X-View(\%) & X-Sub(\%) & X-Set(\%)\\
\hline
J                   & 90.9	&95.9	&86.0	&88.2\\
B	                & 91.2	&95.7	&88.1	&89.2\\
JM	        & 88.5	&94.0	&82.7	&84.6\\
BM	        & 88.1	&93.3	&81.8	&84.2\\
\hline
J+B+JM+BM	& 93.1	&97.1	&90.1	&92.5\\
\hline
\end{tabular}
}
\end{table}

\section{Experiments}
\subsection{Datasets}
We evaluate our method on three widely used skeleton-based action recognition datasets, NTU
RGB+D \cite{shahroudy2016ntu}, NTU RGB+D 120 \cite{liu2019ntu}, and NW-UCLA \cite{wang2014cross}

\subsubsection{NTU RGB+D} \cite{shahroudy2016ntu} is a widely used skeleton-based human action recognition dataset that contains 56,880 skeletal sequences and over 60 action classes. The dataset provides 3D Cartesian coordinates of 25 joints for each human in an action sample, which are captured from three Microsoft Kinect v2 cameras with different viewpoints. Each action sample is performed by 40 volunteers from different age groups. The authors recommend two evaluation benchmarks: (1) \textit{Cross-Subject} (X-Sub), where the 40 subjects are divided into training and testing groups, and (2) \textit{Cross-View} (X-View), where data from cameras 2 and 3 are used for training, and data from camera view 1 is used for testing.

\subsubsection{NTU RGB+D 120} \cite{liu2019ntu} is an extended version of NTU RGB+D. It includes an additional 57,367 skeleton sequences across 60 new action classes.
Similar to the original dataset, two evaluation benchmarks are recommended: (1) \textit{Cross-Subject} (X-Sub), where the 106 subjects are divided into training and testing groups, and (2) \textit{Cross-Setup} (X-Set), where data from samples with even setup IDs are used for training, and data from samples with odd setup IDs are used for testing.

\subsubsection{NW-UCLA} \cite{wang2014cross} is a human action recognition dataset consisting of 1,494 sequences across 10 action classes, captured from three Kinect cameras.

\subsection{Implementation Details}
Our extensive experiments are conducted on an A100 GPU using the PyTorch framework. We train our model for a total of 120 epochs, applying Stochastic Gradient Descent (SGD) with a momentum of 0.9 and a weight decay of 0.0004 for optimization. To stabilize the training process, we use the warmup strategy from\cite{he2016deep} for the first 5 epochs. The initial learning rate is set to 0.1 and is reduced by a factor of 0.1 at the 60th and 90th epochs. Additionally, label smoothing \cite{szegedy2016rethinking} with a weight of 0.1 is adopted to improve generalization.

For NTU RGB+D and NTU RGB+D 120, we set the batch size to 128, and each sample is resized to 64 frames using linear interpolation, as described in \cite{chen2021channel}. For the NW-UCLA dataset, a batch size of 32 is used.

\begin{table*}[t]
\centering
\caption{Top-1 accuracy comparison results of individual experiments and pairwise combination experiments with different hypergraph topologies are presented.  J stands for Joint, B stands for Bone, JM and BM stand for Joint motion and Bone motion.
\label{tab:table4}}
\begin{tabular}{cccccccc}
\hline
\multirow{2}{*}{Tepology} & \multirow{2}{*}{Param(M)} &\multirow{2}{*}{GFLOPS(G)} & \multicolumn{5}{c}{Acc(\%)} \\ \cline{4-8}
 & & &J &B &JM &BM &J+B+JM+BM \\
\hline
h1            &1.08	&1.34	&85.64 &86.85 &81.47 &81.36 &89.45 \\
h2		      &1.08	&1.34   &85.65 &86.96 &81.67 &81.60 &89.60 \\
h3            &1.08	&1.34	&85.74 &87.00 &81.44 &81.43 &89.74 \\
h1 + h2       &1.44	&1.80   &85.82 &87.57 &81.85 &81.88 &89.89\\
h1 + h3       &1.44	&1.80   &86.04 &87.52 &82.08 &81.87 &89.92 \\
h2 + h3       &1.44	&1.80   &86.13 &87.25 &82.31 &81.91 &89.86 \\
h1 + h2 + h3  &1.81	&2.24	&86.04 &88.09 &82.70 &81.82 &\textbf{90.09} \\
\hline
\end{tabular}
\end{table*}

\begin{table}[t]
\centering
\caption{top-1 accuracy comparison with different components on NTU RGB+D 120 X-sub. “+” denotes maintaining the current setting and adding more.
\label{tab:table3}}
\resizebox{1.0\linewidth}{!}{
\begin{tabular}{ccccc}
\hline
Method & Param(M) & GFLOPS(G) &Acc(\%) \\
\hline
baseline	          &1.18	&1.46	&84.1\\
baseline + GCM		  &1.46	&1.97   &84.9\\
baseline + HGCM	      &1.63	&2.03	&85.4\\
baseline + HGCM + AM	  &1.58	&1.98	&85.5\\
baseline + HGCM + HAM  &1.81	&2.24	&86.0\\
\hline
\end{tabular}
}
\end{table}

\subsection{Comparison With the State-of-the-Art Methods}
In this subsection, we compare our proposed method with a comprehensive list of state-of-the-art skeleton-based action recognition methods across all three datasets.  We adopt a multi-stream fusion strategy and utilize four different modalities: joint, bone, joint motion, and bone motion. These modalities are explained as follows: The joint modality refers to the original skeleton coordinates, the bone modality represents the differential of spatial coordinates, and the joint motion and bone motion modalities capture the differential along the temporal dimension of the joint and bone modalities, respectively. The results from these four modalities are fused to produce the final outcome.

Table \ref{tab:table1} presents the results of our comparative experiments. As shown in the table, the recognition performance of our method outperforms existing methods on both the NTU RGB+D and NTU RGB+D 120 benchmarks. Specifically, compared to ST-GCN, our model achieves a 9.4\% and 7.9\% improvement on the NTU RGB+D 60 dataset across the two benchmarks, and a 14.3\% and 16\% improvement on the NTU RGB+D 120 dataset.

When compared to the current state-of-the-art method, InfoGCN, our approach surpasses InfoGCN in all four benchmarks on both NTU RGB+D 60 and NTU RGB+D 120 datasets, with improvements of 0.4\%, 0.2\%, 0.7\%, and 0.8\%, respectively. This demonstrates that our model significantly improves accuracy over other mainstream methods on these two datasets.

For NW-UCLA, our proposed method achieved an accuracy of 96.1\%,  slightly worse than 96.6\% of InfoGCN. This dataset presents particular challenges, as it contains fewer training samples, which makes it harder for our model to perform optimally. Furthermore, the dataset only includes 20 skeleton points, which restricts the effective utilization of hypergraph topology.

Table \ref{tab:table2} shows the accuracy of our model across four different modalities and four test benchmarks from two datasets. 
The results clearly show that using bone information is better than using either joint or motion information alone. Additionally, the inclusion of motion information further improves performance in all scenarios.
This is because skeleton information encompasses more hierarchical features and is more suitable for directed topological modeling.
The fusion of multiple modalities leads to the best performance. This result further validates the superiority of our method.

\subsection{Ablation Studies}
In this subsection, we investigate the impact of different modules in our method on its overall performance. We use ST-GCN\cite{yan2018spatial} as the baseline and analyze the contributions of individual modules by sequentially adding them. For the ablation study, all experiments are conducted on the X-Sub benchmark of NTU RGB+D 120. Only the joint modality is used as input for these experiments.

\subsubsection{Hypergraph Topology Studies}
For the hypergraph topology we designed, we investigated whether different hypergraph topologies might yield varying accuracy in identifying different actions. We conduct experiments on three distinct hypergraph topologies and their pairwise combinations. In addition, we compare the model parameter size and recognition accuracy across different configurations. These configurations include the fusion of data from different modalities.
The experimental results are presented in Table \ref{tab:table4}.

As shown in the table, the model that integrates all three hypergraph topologies achieves the best recognition performance. Among the individual topologies, h3 demonstrates the highest accuracy when all four modalities are fused. For the two-topology combinations, the pairing of h1 and h3 achieves the best accuracy. Nonetheless, the highest recognition accuracy is consistently obtained when all three hypergraph topologies are combined.

\subsubsection{Hypergraph Module Studies}
To verify the effectiveness of the hypergraph module and the contribution of different components in our method, we compare our hypergraph module with configurations that do not include the hypergraph component.
And we analyze the size and accuracy of our model with and without the hypergraph topology. 
The comparison includes four configurations: the Graph convolution module with hypergraph (HGCM) and without hypergraph(GCM), as well as the Attention module with hypergraph (HAM) and without hypergraph (AM).

As shown in the Table \ref{tab:table3}, the recognition performance of model improves progressively with the addition of components. The results demonstrate that incorporating the hypergraph module in our method significantly enhances the baseline recognition performance of model. Specifically, the inclusion of our proposed HGCM and HAM modules improves accuracy by approximately 1.9\%. Although the hypergraph structure increases the parameter count of model, the overall model size remains compact, and the accuracy gains are substantial.

Notably, our HAM and HGCM capture the high-order relationships between different skeleton points. This approach highlights the advantages of using high-order relationships rather than point pairs in topological modeling.
These results validate the effectiveness of our method. By leveraging hypergraph topology for high-order modeling between skeleton points, our approach leads to significant improvements in performance.



\section{Conclusion}
In this paper, we propose a Hypergraph Fused Graph Convolutional Network (HFGCN) for action recognition. We design different skeleton point classification strategies tailored to human motion characteristics to construct hypergraph topologies. These strategies enable the model to focus on the correlations within skeleton point groups. We further employ a Hypergraph Attention Module (HAM) to learn high-order relations between skeleton points, optimizing the topological modeling at the temporal frame level. Additionally, we introduce a Hypergraph Convolution Module (HGCM) to refine the correlations between skeleton points and skeleton point groups at the spatial channel level.
Experimental results on the NTU RGB+D and NTU RGB+D 120 datasets demonstrate that the proposed HFGCN outperforms existing skeleton-based methods. However, the hypergraph in our model is manually designed based on human motion theory. Future work should focus on optimizing this aspect to achieve further improvements.



\bibliographystyle{named}
\bibliography{ijcai25}

\begin{thebibliography}{}

\bibitem[\protect\citeauthoryear{Bawa \bgroup \em et al.\egroup }{2021}]{bawa2021review}
Anthony Bawa, Konstantinos Banitsas, and Maysam Abbod.
\newblock A review on the use of microsoft kinect for gait abnormality and postural disorder assessment.
\newblock {\em Journal of Healthcare Engineering}, 2021(1):4360122, 2021.

\bibitem[\protect\citeauthoryear{Bretto}{2013}]{bretto2013hypergraph}
Alain Bretto.
\newblock Hypergraph theory.
\newblock {\em An introduction. Mathematical Engineering. Cham: Springer}, 1, 2013.

\bibitem[\protect\citeauthoryear{Chen \bgroup \em et al.\egroup }{2021}]{chen2021channel}
Yuxin Chen, Ziqi Zhang, Chunfeng Yuan, Bing Li, Ying Deng, and Weiming Hu.
\newblock Channel-wise topology refinement graph convolution for skeleton-based action recognition.
\newblock In {\em Proceedings of the IEEE/CVF international conference on computer vision}, pages 13359--13368, 2021.

\bibitem[\protect\citeauthoryear{Chen \bgroup \em et al.\egroup }{2022}]{chen2022informed}
Yanjun Chen, Ying Li, Chongyang Zhang, Hao Zhou, Yan Luo, and Chuanping Hu.
\newblock Informed patch enhanced hypergcn for skeleton-based action recognition.
\newblock {\em Information Processing \& Management}, 59(4):102950, 2022.

\bibitem[\protect\citeauthoryear{Cheng \bgroup \em et al.\egroup }{2020}]{cheng2020skeleton}
Ke~Cheng, Yifan Zhang, Xiangyu He, Weihan Chen, Jian Cheng, and Hanqing Lu.
\newblock Skeleton-based action recognition with shift graph convolutional network.
\newblock In {\em Proceedings of the IEEE/CVF conference on computer vision and pattern recognition}, pages 183--192, 2020.

\bibitem[\protect\citeauthoryear{Chi \bgroup \em et al.\egroup }{2022}]{chi2022infogcn}
Hyung-gun Chi, Myoung~Hoon Ha, Seunggeun Chi, Sang~Wan Lee, Qixing Huang, and Karthik Ramani.
\newblock Infogcn: Representation learning for human skeleton-based action recognition.
\newblock In {\em Proceedings of the IEEE/CVF conference on computer vision and pattern recognition}, pages 20186--20196, 2022.

\bibitem[\protect\citeauthoryear{Ding \bgroup \em et al.\egroup }{2019}]{ding2019attention}
Xiaolu Ding, Kai Yang, and Wai Chen.
\newblock An attention-enhanced recurrent graph convolutional network for skeleton-based action recognition.
\newblock In {\em Proceedings of the 2019 2nd International Conference on Signal Processing and Machine Learning}, pages 79--84, 2019.

\bibitem[\protect\citeauthoryear{Du \bgroup \em et al.\egroup }{2015}]{du2015hierarchical}
Yong Du, Wei Wang, and Liang Wang.
\newblock Hierarchical recurrent neural network for skeleton based action recognition.
\newblock In {\em Proceedings of the IEEE conference on computer vision and pattern recognition}, pages 1110--1118, 2015.

\bibitem[\protect\citeauthoryear{Feng \bgroup \em et al.\egroup }{2019}]{feng2019hypergraph}
Yifan Feng, Haoxuan You, Zizhao Zhang, Rongrong Ji, and Yue Gao.
\newblock Hypergraph neural networks.
\newblock In {\em Proceedings of the AAAI conference on artificial intelligence}, volume~33, pages 3558--3565, 2019.

\bibitem[\protect\citeauthoryear{Hao \bgroup \em et al.\egroup }{2021}]{hao2021hypergraph}
Xiaoke Hao, Jie Li, Yingchun Guo, Tao Jiang, and Ming Yu.
\newblock Hypergraph neural network for skeleton-based action recognition.
\newblock {\em IEEE Transactions on Image Processing}, 30:2263--2275, 2021.

\bibitem[\protect\citeauthoryear{He \bgroup \em et al.\egroup }{2016}]{he2016deep}
Kaiming He, Xiangyu Zhang, Shaoqing Ren, and Jian Sun.
\newblock Deep residual learning for image recognition.
\newblock In {\em Proceedings of the IEEE conference on computer vision and pattern recognition}, pages 770--778, 2016.

\bibitem[\protect\citeauthoryear{Host and Iva{\v{s}}i{\'c}-Kos}{2022}]{host2022overview}
Kristina Host and Marina Iva{\v{s}}i{\'c}-Kos.
\newblock An overview of human action recognition in sports based on computer vision.
\newblock {\em Heliyon}, 8(6), 2022.

\bibitem[\protect\citeauthoryear{Huang \bgroup \em et al.\egroup }{2020}]{huang2020part}
Linjiang Huang, Yan Huang, Wanli Ouyang, and Liang Wang.
\newblock Part-level graph convolutional network for skeleton-based action recognition.
\newblock In {\em Proceedings of the AAAI conference on artificial intelligence}, volume~34, pages 11045--11052, 2020.

\bibitem[\protect\citeauthoryear{Kipf and Welling}{2017}]{kipf2017semisupervised}
Thomas~N. Kipf and Max Welling.
\newblock Semi-supervised classification with graph convolutional networks.
\newblock In {\em International Conference on Learning Representations}, 2017.

\bibitem[\protect\citeauthoryear{Li \bgroup \em et al.\egroup }{2019}]{li2019actional}
Maosen Li, Siheng Chen, Xu~Chen, Ya~Zhang, Yanfeng Wang, and Qi~Tian.
\newblock Actional-structural graph convolutional networks for skeleton-based action recognition.
\newblock In {\em Proceedings of the IEEE/CVF conference on computer vision and pattern recognition}, pages 3595--3603, 2019.

\bibitem[\protect\citeauthoryear{Liu \bgroup \em et al.\egroup }{2019}]{liu2019ntu}
Jun Liu, Amir Shahroudy, Mauricio Perez, Gang Wang, Ling-Yu Duan, and Alex~C Kot.
\newblock Ntu rgb+ d 120: A large-scale benchmark for 3d human activity understanding.
\newblock {\em IEEE transactions on pattern analysis and machine intelligence}, 42(10):2684--2701, 2019.

\bibitem[\protect\citeauthoryear{Liu \bgroup \em et al.\egroup }{2020}]{liu2020disentangling}
Ziyu Liu, Hongwen Zhang, Zhenghao Chen, Zhiyong Wang, and Wanli Ouyang.
\newblock Disentangling and unifying graph convolutions for skeleton-based action recognition.
\newblock In {\em Proceedings of the IEEE/CVF conference on computer vision and pattern recognition}, pages 143--152, 2020.

\bibitem[\protect\citeauthoryear{Liu \bgroup \em et al.\egroup }{2023}]{liu2023temporal}
Jinfu Liu, Xinshun Wang, Can Wang, Yuan Gao, and Mengyuan Liu.
\newblock Temporal decoupling graph convolutional network for skeleton-based gesture recognition.
\newblock {\em IEEE Transactions on Multimedia}, 26:811--823, 2023.

\bibitem[\protect\citeauthoryear{Pareek and Thakkar}{2021}]{pareek2021survey}
Preksha Pareek and Ankit Thakkar.
\newblock A survey on video-based human action recognition: recent updates, datasets, challenges, and applications.
\newblock {\em Artificial Intelligence Review}, 54(3):2259--2322, 2021.

\bibitem[\protect\citeauthoryear{Poppe}{2010}]{poppe2010survey}
Ronald Poppe.
\newblock A survey on vision-based human action recognition.
\newblock {\em Image and vision computing}, 28(6):976--990, 2010.

\bibitem[\protect\citeauthoryear{Shahroudy \bgroup \em et al.\egroup }{2016}]{shahroudy2016ntu}
Amir Shahroudy, Jun Liu, Tian-Tsong Ng, and Gang Wang.
\newblock Ntu rgb+ d: A large scale dataset for 3d human activity analysis.
\newblock In {\em Proceedings of the IEEE conference on computer vision and pattern recognition}, pages 1010--1019, 2016.

\bibitem[\protect\citeauthoryear{Shi \bgroup \em et al.\egroup }{2019}]{shi2019two}
Lei Shi, Yifan Zhang, Jian Cheng, and Hanqing Lu.
\newblock Two-stream adaptive graph convolutional networks for skeleton-based action recognition.
\newblock In {\em Proceedings of the IEEE/CVF conference on computer vision and pattern recognition}, pages 12026--12035, 2019.

\bibitem[\protect\citeauthoryear{Simonyan and Zisserman}{2014}]{simonyan2014two}
Karen Simonyan and Andrew Zisserman.
\newblock Two-stream convolutional networks for action recognition in videos.
\newblock {\em Advances in neural information processing systems}, 27, 2014.

\bibitem[\protect\citeauthoryear{Singh \bgroup \em et al.\egroup }{2016}]{singh2016multi}
Bharat Singh, Tim~K Marks, Michael Jones, Oncel Tuzel, and Ming Shao.
\newblock A multi-stream bi-directional recurrent neural network for fine-grained action detection.
\newblock In {\em Proceedings of the IEEE conference on computer vision and pattern recognition}, pages 1961--1970, 2016.

\bibitem[\protect\citeauthoryear{Song \bgroup \em et al.\egroup }{2020}]{song2020stronger}
Yi-Fan Song, Zhang Zhang, Caifeng Shan, and Liang Wang.
\newblock Stronger, faster and more explainable: A graph convolutional baseline for skeleton-based action recognition.
\newblock In {\em proceedings of the 28th ACM international conference on multimedia}, pages 1625--1633, 2020.

\bibitem[\protect\citeauthoryear{Song \bgroup \em et al.\egroup }{2021}]{song2021human}
Liangchen Song, Gang Yu, Junsong Yuan, and Zicheng Liu.
\newblock Human pose estimation and its application to action recognition: A survey.
\newblock {\em Journal of Visual Communication and Image Representation}, 76:103055, 2021.

\bibitem[\protect\citeauthoryear{Szegedy \bgroup \em et al.\egroup }{2016}]{szegedy2016rethinking}
Christian Szegedy, Vincent Vanhoucke, Sergey Ioffe, Jon Shlens, and Zbigniew Wojna.
\newblock Rethinking the inception architecture for computer vision.
\newblock In {\em Proceedings of the IEEE conference on computer vision and pattern recognition}, pages 2818--2826, 2016.

\bibitem[\protect\citeauthoryear{Thakkar and Narayanan}{2018}]{thakkar2018part}
Kalpit Thakkar and PJ~Narayanan.
\newblock Part-based graph convolutional network for action recognition.
\newblock {\em arXiv preprint arXiv:1809.04983}, 2018.

\bibitem[\protect\citeauthoryear{Tran \bgroup \em et al.\egroup }{2015}]{tran2015learning}
Du~Tran, Lubomir Bourdev, Rob Fergus, Lorenzo Torresani, and Manohar Paluri.
\newblock Learning spatiotemporal features with 3d convolutional networks.
\newblock In {\em Proceedings of the IEEE international conference on computer vision}, pages 4489--4497, 2015.

\bibitem[\protect\citeauthoryear{Vemulapalli \bgroup \em et al.\egroup }{2014}]{vemulapalli2014human}
Raviteja Vemulapalli, Felipe Arrate, and Rama Chellappa.
\newblock Human action recognition by representing 3d skeletons as points in a lie group.
\newblock In {\em Proceedings of the IEEE conference on computer vision and pattern recognition}, pages 588--595, 2014.

\bibitem[\protect\citeauthoryear{Wang and Wang}{2017}]{wang2017modeling}
Hongsong Wang and Liang Wang.
\newblock Modeling temporal dynamics and spatial configurations of actions using two-stream recurrent neural networks.
\newblock In {\em Proceedings of the IEEE conference on computer vision and pattern recognition}, pages 499--508, 2017.

\bibitem[\protect\citeauthoryear{Wang \bgroup \em et al.\egroup }{2014}]{wang2014cross}
Jiang Wang, Xiaohan Nie, Yin Xia, Ying Wu, and Song-Chun Zhu.
\newblock Cross-view action modeling, learning and recognition.
\newblock In {\em Proceedings of the IEEE conference on computer vision and pattern recognition}, pages 2649--2656, 2014.

\bibitem[\protect\citeauthoryear{Wu \bgroup \em et al.\egroup }{2022}]{wu2022survey}
Fei Wu, Qingzhong Wang, Jiang Bian, Ning Ding, Feixiang Lu, Jun Cheng, Dejing Dou, and Haoyi Xiong.
\newblock A survey on video action recognition in sports: Datasets, methods and applications.
\newblock {\em IEEE Transactions on Multimedia}, 25:7943--7966, 2022.

\bibitem[\protect\citeauthoryear{Xia \bgroup \em et al.\egroup }{2012}]{xia2012view}
Lu~Xia, Chia-Chih Chen, and Jake~K Aggarwal.
\newblock View invariant human action recognition using histograms of 3d joints.
\newblock In {\em 2012 IEEE computer society conference on computer vision and pattern recognition workshops}, pages 20--27. IEEE, 2012.

\bibitem[\protect\citeauthoryear{Yan \bgroup \em et al.\egroup }{2018}]{yan2018spatial}
Sijie Yan, Yuanjun Xiong, and Dahua Lin.
\newblock Spatial temporal graph convolutional networks for skeleton-based action recognition.
\newblock In {\em Proceedings of the AAAI conference on artificial intelligence}, volume~32, 2018.

\bibitem[\protect\citeauthoryear{Yang \bgroup \em et al.\egroup }{2019}]{yang2019gesture}
LI~Yang, Jin Huang, TIAN Feng, WANG Hong-An, and DAI Guo-Zhong.
\newblock Gesture interaction in virtual reality.
\newblock {\em Virtual Reality \& Intelligent Hardware}, 1(1):84--112, 2019.

\bibitem[\protect\citeauthoryear{Zhou \bgroup \em et al.\egroup }{2022}]{zhou2022hypergraph}
Yuxuan Zhou, Zhi-Qi Cheng, Chao Li, Yanwen Fang, Yifeng Geng, Xuansong Xie, and Margret Keuper.
\newblock Hypergraph transformer for skeleton-based action recognition.
\newblock {\em arXiv preprint arXiv:2211.09590}, 2022.

\end{thebibliography}

\end{document}